\documentclass{article}
\usepackage{arxiv}
\usepackage{amsmath,amssymb,amsfonts}
\usepackage{graphicx}
\usepackage[export]{adjustbox}
\usepackage{textcomp}
\usepackage{xcolor}
\usepackage{biblatex}
\usepackage{booktabs}
\usepackage{multirow}
\usepackage{subcaption}
\usepackage{svg}
\usepackage{bm}
\usepackage{hyperref}
\usepackage{cleveref}
\usepackage{bm}

\usepackage{algorithm}
\usepackage{algorithmicx}
\usepackage{algpseudocode}

\renewcommand{\shorttitle}{Diffusion Crossover}

\title{Diffusion Crossover: Defining Evolutionary Recombination in Diffusion Models via Noise Sequence Interpolation}

\author{ 
{Chisato Kumada}\\
Graduate School of Life and Medical Sciences\\
\And
{Satoru Hiwa}\\
Faculty of Life and Medical Sciences\\
Doshisha University\\
\And
{Tomoyuki Hiroyasu}\\
Faculty of Life and Medical Sciences\\
Doshisha University\\
Kyoto, Japan\\
\texttt{tomo@is.doshisha.ac.jp} 
}

\begin{document}
\maketitle

\begin{abstract}
Interactive Evolutionary Computation (IEC) provides a powerful framework for optimizing subjective criteria such as human preferences and aesthetics, yet it suffers from a fundamental limitation: in high-dimensional generative representations, defining crossover in a semantically consistent manner is difficult, often leading to a mutation-dominated search.
In this work, we explicitly define crossover in diffusion models.
We propose \emph{Diffusion crossover}, which formulates evolutionary recombination as step-wise interpolation of noise sequences in the reverse process of Denoising Diffusion Probabilistic Models (DDPMs).
By applying spherical linear interpolation (Slerp) to the noise sequences associated with selected parent images, the proposed method generates offspring that inherit characteristics from both parents while preserving the geometric structure of the diffusion process.
Furthermore, controlling the time-step range of interpolation enables a principled trade-off between diversity (exploration) and convergence (exploitation).
Experimental results using PCA analysis and perceptual similarity metrics (LPIPS) demonstrate that Diffusion crossover produces perceptually smooth and semantically consistent transitions between parent images.
Qualitative interactive evolution experiments further confirm that the proposed method effectively supports human-in-the-loop image exploration.
These findings suggest a new perspective: diffusion models are not only powerful generators, but also structured evolutionary search spaces in which recombination can be explicitly defined and controlled.
\end{abstract}

Diffusion Models, 
Denoising Diffusion Probabilistic Models (DDPM), 
Interactive Evolutionary Computation (IEC), 
Diffusion crossover, 
Noise Sequence Interpolation, 
Human-in-the-loop Optimization

\section{Introduction}
With the advancement of information and communication technology (ICT), particularly generative AI, it has become feasible to generate designs aligned with users' preferences and sensibilities.
However, personal ``preferences'' and ``affective judgments'' are difficult to formalize as explicit mathematical functions,
posing challenges for conventional optimization and automated generation methods.

Interactive Evolutionary Computation (IEC) \cite{takagi2001interactive} provides an effective framework to address this issue.
Unlike conventional evolutionary computation, which relies on predefined fitness functions,
IEC allows humans to directly evaluate individuals based on subjective criteria.
By iterating candidate generation and human evaluation,
IEC enables efficient exploration of subjectively optimal solutions.

The IEC has been applied to domains that require affective evaluation, such as kimono design \cite{sugahara2008design} and furniture design \cite{Sawada2025evolutionary}.
However, the quality and diversity of generated individuals strongly affect search efficiency and user burden.
Low-quality outputs hinder evaluation, while insufficient diversity leads to premature convergence.
In high-dimensional design spaces, designing crossover operators that preserve semantics while introducing novelty is difficult,
often resulting in mutation-centered exploration.

To address these issues, recent studies have incorporated generative models into IEC \cite{bontrager2018deep}.
Among them, diffusion models have attracted attention for their high generation performance and stable training \cite{dhariwal2021diffusion}.
Denoising Diffusion Probabilistic Models (DDPM) \cite{ho2020denoising} generate data via a forward noise addition process and a reverse denoising process.
By modeling noise prediction, DDPM achieves high-quality generation.
This process forms a coarse-to-fine generation trajectory,
making diffusion models suitable for controllable and diverse generation in IEC.

Recent studies have clarified the relationship between diffusion models and evolutionary computation.
Zhang et al. \cite{zhang2025diffusion} showed that mutation, selection, and inter-individual recombination are embedded in diffusion processes and proposed a Diffusion Evolution algorithm for numerical optimization.
These findings suggest that diffusion models can be interpreted as evolutionary search dynamics.

However, existing studies mainly focus on numerical optimization,
and the application to IEC remains limited, especially for explicit crossover design.
Diffusion models assume one output per noise sample,
and integrating information from multiple parents is not straightforward.
Thus, new crossover operators grounded in the generation process are required.

In this study, we address this problem by focusing on noise components and time steps in the reverse diffusion process, and define a novel \emph{Diffusion crossover} based on noise sequence interpolation.
The main contributions are:
1) We formulate crossover operations in IEC within the diffusion generation process and propose Diffusion crossover in noise space.
2) Leveraging DDPM ~\cite{ho2020denoising} and theoretical insights by Zhang et al. ~\cite{zhang2025diffusion}, we extend diffusion models to IEC with human evaluation.
3) We demonstrate that Diffusion crossover improves search efficiency in terms of quality and diversity, showing strong potential to reduce user burden in human-in-the-loop exploration.

Through this approach, we enable evolutionary exploration that reflects human preferences
while preserving the generation capability of diffusion models.

\section{Background}
\subsection{Evolutionary Generative Design with Deep Generative Models}
With the advancement of deep generative models, recent studies have explored
Interactive Genetic Algorithms (IGAs) in which the latent variables of generative models
are treated as search spaces.
Early approaches utilized Generative Adversarial Networks (GANs)~\cite{bontrager2018deep}, Variational Autoencoders (VAEs)~\cite{guoindirect2018}, or 3D point cloud autoencoders~\cite{riosmultitask2022},
where latent vectors are regarded as genotypes,
and crossover is realized through interpolation or arithmetic operations between vectors.
However, GANs suffer from inherent issues such as training instability, and autoencoder-based models often struggle to achieve high-quality image synthesis.
In contrast, diffusion models exhibit stable training characteristics
and the ability to generate high-quality and diverse images,
making them a more suitable foundation for Evolutionary Generative Design.

\subsection{Diffusion and Denoising Process of Diffusion Models}
Diffusion models formulate the image generation process
as a gradual transition from noise to the target data distribution.
Due to their stable training behavior and high-quality generation capability,
they have been applied in various domains,
including image synthesis, audio generation, and medical imaging.
In this study, we adopt DDPM~\cite{ho2020denoising}
as a representative diffusion-based generative model.

DDPM consists of two main processes:
a forward diffusion process that progressively adds noise to images,
and a reverse diffusion (denoising) process that removes noise step by step.
DDPM serves as the foundation for many subsequent diffusion-based models.
We choose DDPM not only because it is a canonical diffusion model,
but also because its stochastic sampling process plays an essential role
in preserving diversity in the generated images within our framework.

In the forward diffusion process,
Gaussian noise is gradually added to a training image $\bm{x}_0$,
eventually transforming it into a pure noise image $\bm{x}_T$.
Although noise is added sequentially from $t=1$ to $t=T$, the Gaussian property allows $\bm{x}_t$ at an arbitrary step
to be directly computed from $\bm{x}_0$ as follows:
\begin{equation}
     \bm{x}_t = \sqrt{\overline{\alpha}_t}\bm{x}_0 + \sqrt{1-\overline{\alpha}_t}\epsilon,
     \label{eq:alpha_x0}
\end{equation}
where $\alpha_t = 1 - \beta_t$,
$\overline{\alpha}_t = \prod_{s=1}^t \alpha_s$,
and $\epsilon \sim \mathcal{N}(\bm{0}, \bm{I})$.
This reparameterization is used during the diffusion model's training.

In the reverse diffusion process,
images are gradually reconstructed from noise.
A U-Net-based neural network $\epsilon_\theta$~\cite{ronneberger2015unetconvolutionalnetworksbiomedical}
is trained to predict the noise component contained in $\bm{x}_t$.
Sampling using the trained model is performed sequentially according to:
\begin{gather}
    \bm{\mu}_{\bm{\theta}}(\bm{x}_t, t)
    = \frac{1}{\sqrt{\alpha_t}}
    \left(
        \bm{x}_t
        - \frac{1-\alpha_t}{\sqrt{1-\overline{\alpha}_t}}
        \epsilon_{\bm{\theta}}(\bm{x}_t, t)
    \right), \\
    \bm{x}_{t-1} = \bm{\mu}_{\bm{\theta}}(\bm{x}_t, t) + \sigma_t \bm{z},
    \label{eq:inferrence}
\end{gather}
where $\sigma_t = \sqrt{\beta_t}$
and $\bm{z} \sim \mathcal{N}(\bm{0}, \bm{I})$.
Here, the noise term $\bm{z}$ at each step introduces stochasticity into the generation process,
enabling the generation of diverse images.
In this study, the network $\epsilon_\theta$ is trained
using a squared error loss function, following the formulation by Ho et al.~\cite{ho2020denoising}.

\subsection{Relationship between Diffusion Models and Evolutionary Computation}
Recently, attempts have been made to apply diffusion models
within the framework of evolutionary computation.

From a theoretical perspective,
Diffusion Evolution~\cite{zhang2025diffusion}
interprets the sequence of denoising operations in diffusion models
as an evolutionary process, where the random noise added during reverse diffusion is regarded as a mutation operator.
This interpretation provides a mathematical correspondence
between diffusion processes and evolutionary dynamics.

From a methodological perspective,
diffusion models have also been used as phenotype generators,
mapping latent representations to observable solutions.
However, in existing studies,
implementations of crossover operations remain limited.
Kobayashi et al.~\cite{Kobayashi2023image}
proposed a crossover mechanism based on mixing text prompts and seeds.
However, this approach merely interpolates the initial conditions and does not operate on the generative process itself.
Voronoi Crossover~\cite{mathurin2023interactive}
performs spatial partitioning in the latent space
and generates offspring by cutting and recombining latent variables of parent images.
This method, however, requires explicit region specification by the user
and differs from semantic-level integration
that fuses global styles or image structures.
Sobania et al.~\cite{sobania2025image}
explored pixel-space blending and linear or geometric operations
(such as splitting and merging) on latent variables, but their primary goal is optimizing generated images with respect to prompts,
and changes in image features induced by crossover
are not directly analyzed.

In contrast, this study proposes a novel formulation in which crossover in diffusion models
is explicitly defined as the interpolation of noise sequences.
Unlike conventional latent crossovers that only manipulate the initial seed, our approach intervenes in the generation trajectory.
Furthermore, by adjusting the time steps at which interpolation is applied, our method enables explicit control over the diversity of generated individuals and their resemblance to the parents.
By introducing this interpretation, we expand the scope of evolutionary generative operations in diffusion models
and enable more flexible and diversity-aware generation of individuals.

\section{Proposed Method}
\begin{figure*}[t]
\centering
    \includegraphics[
    width=\columnwidth,
      bb=0 0 1700 480,
  clip
  ]{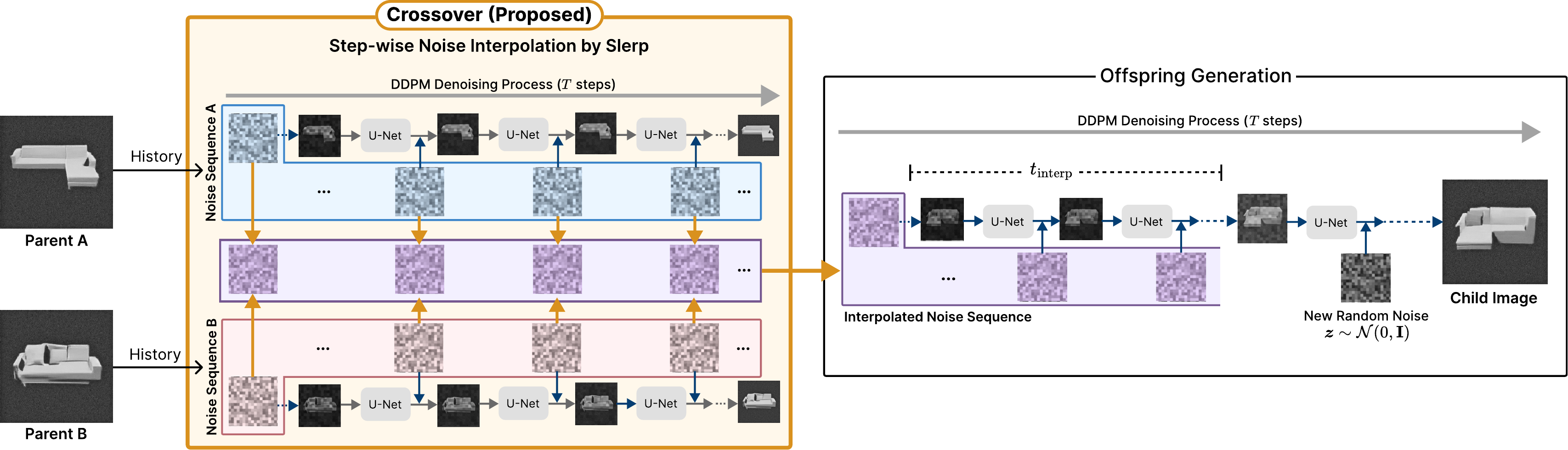}
    \caption{Overview of the proposed Diffusion crossover framework.}
    \label{img:crossover}
\end{figure*}

In this study, we propose a novel framework that integrates
an IGA into the image generation process of DDPM,
enabling image exploration guided by human perceptual preferences (Fig.~\ref{img:crossover}).
The key feature of the proposed method lies in explicitly defining
the crossover operation in evolutionary computation
by focusing on the noise sequence used in the reverse diffusion process.


Conventional evolutionary search methods based on diffusion models primarily emphasize mutation-like operations or the recombination of static inputs, such as prompts or initial latent variables.
In contrast, our approach realizes crossover
by interpolating the noise sequences associated with multiple generated images,
thereby integrating features of parent individuals.
Users select preferred parent images from a set of candidates,
and the system generates the next generation of images
by exploiting the corresponding noise sequences.
Through this human-in-the-loop exploration process,
the proposed framework enables image search 
that directly reflects user intent.

\subsection{Genotype Definition and Crossover via Slerp} \label{sec:crossover}

In the proposed method, the genotype is defined as
the initial noise $\bm{x}_T$ and the sequence of additive noise vectors
used in the reverse diffusion process,
$Z = \{\bm{z}_1, \bm{z}_2, \dots, \bm{z}_T\}$,
where $T$ denotes the maximum number of diffusion time steps.
Crucially, the noise targeted for interpolation in this study
corresponds to the Gaussian noise that is independently sampled (i.e., $\bm{z}_t \sim \mathcal{N}(\bm{0}, \bm{I})$) and added
after denoising at each reverse diffusion step, rather than the noise predicted by the trained network.


Let $\boldsymbol{Z}^{\rm (A)}$ and $\boldsymbol{Z}^{\rm (B)}$ be the noise sequence corresponding to images selected by the user.
To perform crossover, we apply spherical linear interpolation (Slerp) to these noise elements at each time step $t$.
Since each noise term $\bm{z}_t$ originally has the same spatial dimensions as the image, it is flattened into a one-dimensional vector prior to interpolation.
Slerp is an interpolation technique that transitions between two vectors with constant angular velocity while preserving the norm, making it suitable for manipulating high-dimensional Gaussian noise.
The interpolated noise $\bm{z}^{(\lambda)}$ at each time step $t$ is calculated as follows:
\begin{equation}
\boldsymbol{z}^{(\lambda)}_t = \frac{\sin((1 - \lambda) \theta)}{\sin \theta} \boldsymbol{z}^{\rm (A)}_t + \frac{\sin(\lambda \theta)}{\sin \theta} \boldsymbol{z}^{\rm (B)}_t
\end{equation}
where $\theta = \arccos \left( \boldsymbol{z}^{\rm (A)}_t \cdot \boldsymbol{z}^{\rm (B)}_t / \|\boldsymbol{z}^{\rm (A)}_t\| \|\boldsymbol{z}^{\rm (B)}_t\| \right)$ is the angle between the two vectors and $\lambda$ is the interpolation coefficient.

\subsection{Exploration and Exploitation Control}

To realize effective image exploration, we divide the reverse diffusion process into two stages controlled by the parameter $t_{\rm interp}$.
In the first stage (from step $T$ to $T - t_{\rm interp}$), we apply the crossover operation using Slerp, as introduced in Section \ref{sec:crossover}.
The interpolated noise $\bm{z}_t^{(\lambda)}$ replaces the standard Gaussian noise term $\bm{z} \sim \mathcal{N}(\bm{0}, \bm{I})$ in the reverse diffusion process. 
In the remaining steps, we stop using interpolated noise and instead sample new Gaussian noise.
This stochastic noise addition serves a function equivalent to mutation in evolutionary computation.

The interpolation duration $t_{\rm interp}$ allows us to manage the balance between exploration and exploitation.
Specifically, increasing $t_{\rm interp}$ gradually as the search progresses causes the generated images to reflect the parents' characteristics more faithfully, guiding the search toward exploitation.
Thus, our method utilizes the temporal structure of diffusion models to explicitly separate crossover and mutation, enabling search control based on user intent.

\subsection{Algorithm}
The procedure for exploring images with the proposed method is shown in Algorithm \ref{alg:gen} and Algorithm \ref{alg:main}.
Algorithm \ref{alg:gen} represents the single image generation process with interpolated noise and Algorithm \ref{alg:main} corresponds to the overall IGA process.

\noindent \textbf{Notation:} $\boldsymbol{X}=\{\bm{x}^{(i)}\}_{i=1}^N$ denotes the population of images with size $N$.
$\bm{x}_t$ represents the image at timestep $t$.
$\bm{Z}=\{\bm{z}_t\}_{t=1}^T$ is the noise sequence added at each step $t$.
$\bm{\epsilon}_\theta$ is the noise prediction network, and $\alpha_t, \bar{\alpha}_t, \sigma_t$ are the diffusion scheduling parameters.
$\lambda$, $t_{\rm interp}$, and $s$ denote the interpolation coefficient, duration, and step increment, respectively.


\begin{algorithm}
\caption{Image Generation with Noise Injection} \label{alg:gen}
\begin{algorithmic}[1]
    \State \textbf{function} \textsc{G}($\bm{x}_T, \bm{Z}^{\text{(in)}}, t_{\text{interp}}$)
    \State \quad $\bm{Z}^{\text{(out)}} \gets \emptyset$ \hfill $\triangleright$ Initialize output noise sequence
    \State \quad $\bm{x} \gets \bm{x}_T$
    \State \quad \textbf{for} $t = T, \dots, 1$ \textbf{do}
    \State \quad \quad \textbf{if} $t > T - t_{\text{interp}}$ \textbf{then and} $\bm{Z}^{\text{(in)}} \neq \text{None}$
    \State \quad \quad \quad $z \gets \bm{Z}_t^{\text{(in)}}$ \hfill $\triangleright$ Use injected noise
    \State \quad \quad \textbf{else}
    \State \quad \quad \quad $z \sim \mathcal{N}(\bm{0}, \bm{I})$ \hfill $\triangleright$ Sample new noise
    \State \quad \quad \textbf{end if}
    \State \quad \quad $\bm{Z}^{\text{(out)}} \gets \bm{Z}^{\text{(out)}} \cup \{z\}$ \hfill $\triangleright$ Record used noise
    \State \quad \quad $\bm{x} \gets \frac{1}{\sqrt{\alpha_t}} \left( \bm{x} - \frac{1-\alpha_t}{\sqrt{1-\bar{\alpha}_t}} \bm{\epsilon}_\theta(\bm{x}, t) \right) + \sigma_t z$
    \State \quad \textbf{end for}
    \State \quad \textbf{return} $\bm{x}, \bm{Z}^{\text{(out)}}$
    \State \textbf{end function}
\end{algorithmic}
\end{algorithm}

\begin{algorithm}
\caption{Evolutionary Image Generation via Noise Slerp} \label{alg:main}
\begin{algorithmic}[1]
    \State $\boldsymbol{X}_T \gets \{\bm{x}_T^{(i)} \sim \mathcal{N}(\bm{0}, \bm{I}) \mid i = 1, 2, \dots, N\}$ 
    \State $\boldsymbol{X}_0, \{\bm{Z}^{(i)}\}_{i=1}^N \gets \{ \Call{G}{\bm{x}, \text{None}, 0} \mid \bm{x} \in \boldsymbol{X}_T \}$
    \State \hfill $\triangleright$ Generate the first population
    
    \While{User continues image selection}
        \State Select indices $A, B$ based on user preference
        \State $\bm{x}_T^{\text{(A)}}, \bm{x}_T^{\text{(B)}} \gets \boldsymbol{X}_T^{\text{(A)}}, \boldsymbol{X}_T^{\text{(B)}}$
        \State $\bm{Z}^{\text{(A)}}, \bm{Z}^{\text{(B)}} \gets \bm{Z}^{\text{(A)}}, \bm{Z}^{\text{(B)}}$
        \State $\boldsymbol{X}_{\text{new}} \gets \emptyset, \ \boldsymbol{Z}_{\text{new}} \gets \emptyset$
        
        \For {$i=1, \dots, N$}
            \State Determine $\lambda \sim \mathcal{U}(0, 1)$, $\bm{Z}^{(i)}_{\text{interp}} \gets \emptyset$
            \State $\bm{x}^{(i)}_{T} \gets \text{Slerp}(\bm{x}^{\text{(A)}}_T, \bm{x}^{\text{(B)}}_T, \lambda)$ \hfill $\triangleright$ Slerp $t=T$ images
            
            \For{$t = T, \dots, T - t_{\text{interp}} + 1$}
                \State $\bm{z} \gets \text{Slerp}(\bm{Z}^{\text{(A)}}_t, \bm{Z}^{\text{(B)}}_t, \lambda)$
                \State $\bm{Z}^{(i)}_{\text{interp}} \gets \bm{Z}^{(i)}_{\text{interp}} \cup \{\bm{z}\}$
            \EndFor
            
            \State $\bm{x}^{(i)}_0, \bm{Z}^{(i)} \gets \Call{G}{\bm{x}_T^{(i)}, \bm{Z}^{(i)}_{\text{interp}}, t_{\text{interp}}}$
            \State $\boldsymbol{X}_{\text{new}} \gets \boldsymbol{X}_{\text{new}} \cup \{ \bm{x}^{(i)}_0 \}, \ \boldsymbol{Z}_{\text{new}} \gets \boldsymbol{Z}_{\text{new}} \cup \{ \bm{Z}^{(i)} \}$
        \EndFor
        
        \State Update Population with $\boldsymbol{X}_{\text{new}}, \boldsymbol{Z}_{\text{new}}$
        \State $t_{\text{interp}} \gets t_{\text{interp}} + s$
    \EndWhile
\end{algorithmic}
\end{algorithm}

\section{Experiments}
To evaluate the effectiveness of the proposed Diffusion crossover based on noise sequence interpolation, we conducted a series of numerical experiments
as well as qualitative evaluations.
The objectives of the experiments are as follows:
(i) to analyze the geometric behavior of the diffusion process
and the proposed interpolation operation,
(ii) to verify the perceptual continuity of image generation
enabled by the proposed crossover,
(iii) to evaluate the validity of diversity control
via the interpolation duration,
and (iv) to assess the practical feasibility
of the method as an IEC framework.

\subsection{Datasets}
We used the following two image datasets in our experiments.
To focus on intra-class variability rather than inter-class differences, we used images from a single class per dataset.

\begin{figure}[h]
\centering
\begin{minipage}[t]{0.47\columnwidth}
    \centering
    \includegraphics[
    width=\linewidth,
    bb=00 00 500 400,
      clip
    ]{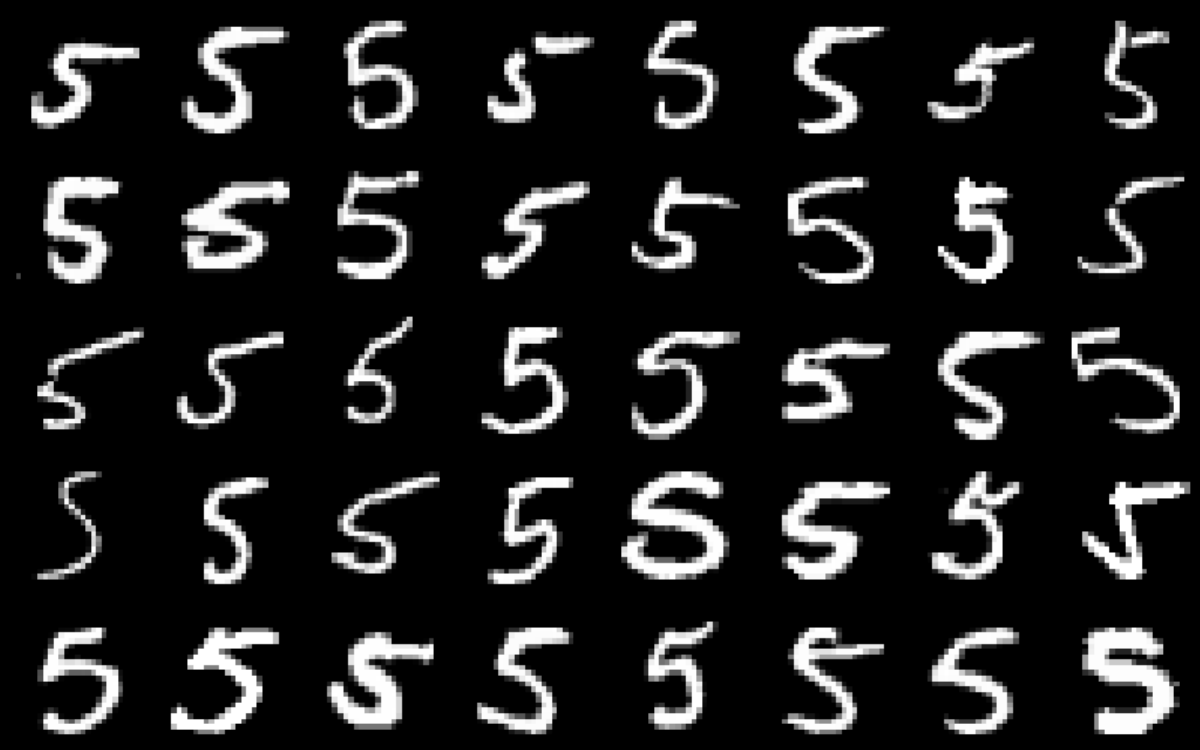}
    \caption{Part of the training dataset (Handwritten ``5'')}
    \label{fig:handwriting_sample}
\end{minipage}
\hfill
\begin{minipage}[t]{0.47\columnwidth}
    \centering
\includegraphics[
width=\linewidth,
bb=00 00 1800 1300,
  clip
    ]{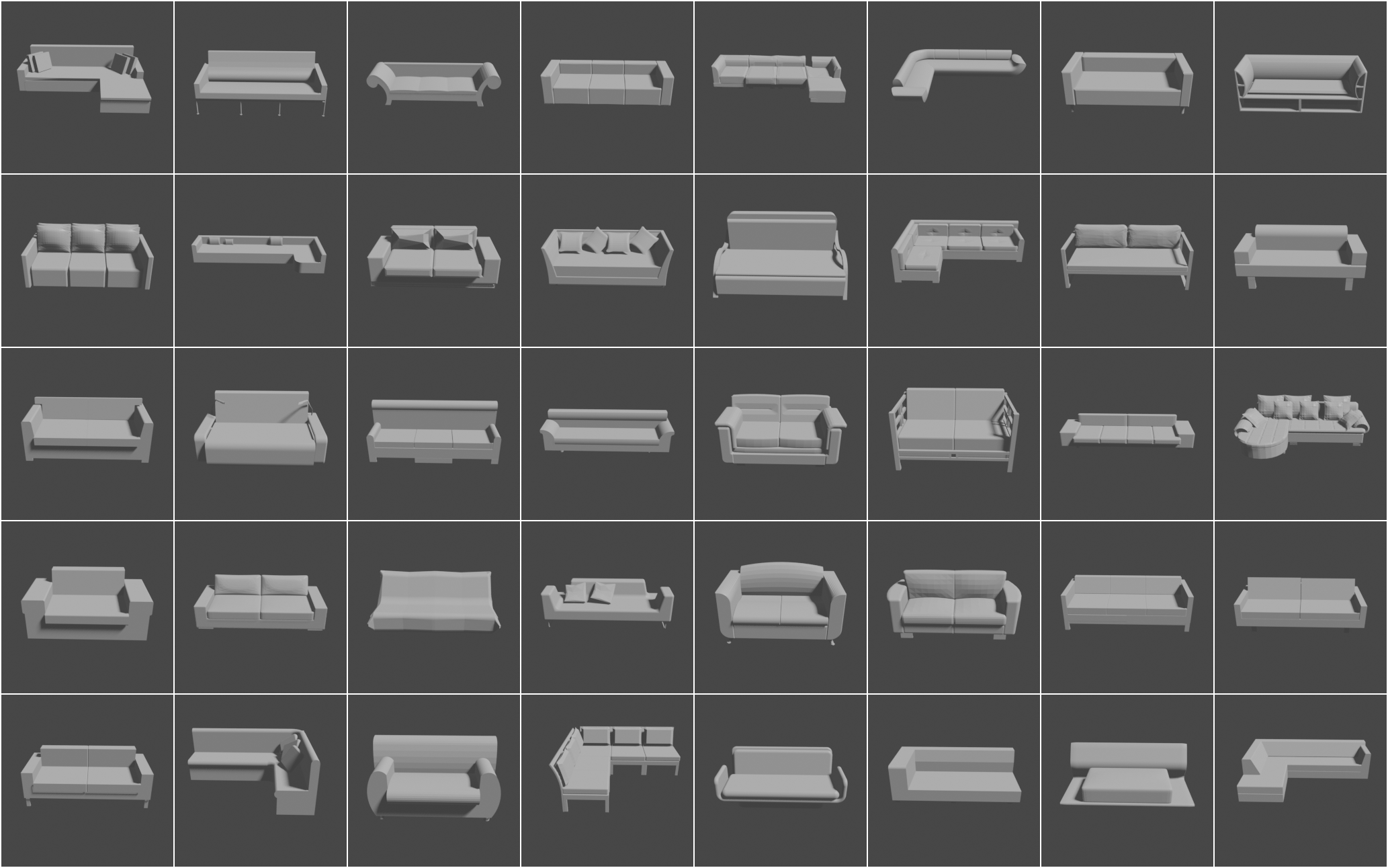}
\caption{Part of the training dataset (Sofa).}
\label{fig:sofa_sample}
\end{minipage}
\end{figure}

\paragraph{MNIST}
From the MNIST dataset, we extracted only images of the digit ``5''.
Each image was resized to $32 \times 32$ pixels
using zero padding.
A subset is shown in Fig.~\ref{fig:handwriting_sample}.

\paragraph{ModelNet}
From the ModelNet40 dataset~\cite{wu20153dshapenets},
we extracted the sofa category
and rendered the 3D shapes into 2D images.
The resolution of the generated images
was set to $256 \times 256$.
A subset is shown in Fig.~\ref{fig:sofa_sample}.

\subsection{Experimental Setup}
We employed DDPM as the diffusion model,
and adopted a U-Net~\cite{ronneberger2015unetconvolutionalnetworksbiomedical}
composed solely of residual blocks
as the generation network.
Since the input images are relatively simple, no attention blocks were used.
Sampling steps were set to $T=1000$,
and a linear noise schedule was adopted.

The models were implemented in PyTorch and optimized using AdamW.
To mitigate overfitting, we applied data augmentation techniques during training.
Specifically, random scaling (0.9 - 1.1), rotation ($\pm$15\%), and translation ($\pm$10\%) were used for MNIST, while random horizontal flipping ($p=0.4$) was applied to ModelNet.
Key hyperparameters were tuned using Optuna~\cite{akiba2019optuna},
while the remaining parameters followed
the default AdamW settings in PyTorch.
Experiments were conducted on an NVIDIA GeForce RTX 4080 Super GPU.
We utilized batch sizes of 64 for MNIST and 40 for ModelNet.
The models were trained for 3{,}000 epochs,
and checkpoints were saved every 100 epochs.
After training,
generated images from each checkpoint
were evaluated using CMMD~\cite{sadeep2024rethinking},
and the model with the lowest CMMD score
was selected for each dataset for subsequent experiments.
The detailed hyperparameter settings
are summarized in Table~\ref{tab:parameters}.

\begin{table}[t]
\centering
\caption{Hyperparameters for training.}
\label{tab:parameters}
\begin{tabular}{@{}lrr@{}}
\toprule
 & \multicolumn{1}{c}{MNIST} & \multicolumn{1}{c}{ModelNet} \\ \midrule 
Training Set   & 5421 & 600 \\
Validation Set & 892  & 80  \\
Batch Size     & 64   & 40  \\
Learning Rate  & $8.6 \times 10^{-4}$ & $5.8 \times 10^{-4}$ \\
$\beta_1$      & $0.73$ & $0.86$ \\
Weight Decay   & $4.2 \times 10^{-3}$ & $5.4 \times 10^{-3}$ \\
\bottomrule
\end{tabular}
\end{table}

\subsection{Experiment 1: Validation of Diffusion crossover Behavior}
In this experiment, we investigate whether the proposed Diffusion crossover
functions as a valid crossover operator
in the context of evolutionary computation.
Specifically, we examine whether
intermediate generation trajectories
are formed between the generation processes
corresponding to parent individuals,
and analyze their behavior.

For each dataset, a separate DDPM was trained.
Starting from random noise $\mathbf{x}_{1000}$,
50 images were generated,
and intermediate images in the reverse diffusion process
were saved every 100 steps, including the final generated image $\bm{x}_0$.
Each saved image was vectorized and normalized,
and Principal Component Analysis (PCA) was applied.
The first and second principal components
were visualized in a two-dimensional space.

In addition, we selected two generated images that were relatively distant in PCA space to serve as parents.
Using these parents, we generated new images from interpolated noise sequences with $t_{\rm interp}=600$, by varying the interpolation coefficient $\lambda$.
For these interpolated processes, intermediate images were also saved every 100 steps, consistent with the standard generation.
The resulting image sequences
were projected into the same space
using the identical normalization and PCA transformation,
allowing direct comparison
with the standard reverse diffusion trajectories.

\subsection{Experiment 2: Continuity of Image Generation via Noise Sequence Interpolation}
To evaluate the effectiveness of the proposed Diffusion crossover, we examined whether noise sequence interpolation
induces perceptually smooth image transitions.

Noise sequences corresponding to randomly selected parent images
$\mathbf{x}^{\rm (A)}$ and $\mathbf{x}^{\rm (B)}$ were used,
and the interpolation coefficient $\lambda$
was varied from $0.1$ to $0.9$ in increments of $0.1$.
The interpolation duration was fixed to $t_{\rm interp}=600$.
For each $\lambda$,
50 images were generated for MNIST
and 10 images for ModelNet,
and their average behavior was used for evaluation.
Five trials were conducted for each dataset.

Perceptual similarity between generated 
and parent images
was evaluated using
Learned Perceptual Image Patch Similarity (LPIPS)~\cite{richard2018unreasonable},
where smaller values indicate higher similarity.
If the interpolation is perceptually smooth,
the LPIPS distance to $\mathbf{x}^{\rm (A)}$
is expected to increase monotonically with $\lambda$,
while the distance to $\mathbf{x}^{\rm (B)}$
is expected to decrease.
To test this hypothesis,
Spearman's rank correlation coefficients
between $\lambda$ and LPIPS values were computed,
and a no-correlation test
was conducted at a significance level of $\alpha=0.05$.

\subsection{Experiment 3: Effect of Interpolation Duration on Diversity}
This experiment evaluates the effect of
the interpolation duration length $t_{\rm interp}$
on the diversity of generated images.

Parent images $\mathbf{x}^{\rm (A)}$ and $\mathbf{x}^{\rm (B)}$ were fixed,
and $t_{\rm interp}$ was varied from 100 to 900
in increments of 100.
The interpolation coefficient was fixed at $\lambda=0.5$.
For each $t_{\rm interp}$,
50 images were generated for MNIST
and 10 images for ModelNet.
Five trials were conducted for each dataset.

Diversity was quantified
as the average pairwise LPIPS
among generated images.
To evaluate whether a monotonic relationship
exists between $t_{\rm interp}$
and the diversity score,
Spearman's rank correlation coefficient was computed,
and a no-correlation test
was performed with $\alpha=0.05$.

\subsection{Experiment 4: Qualitative Evaluation of Interactive Evolutionary Process}
Finally, we qualitatively evaluated the practical feasibility of the proposed method
for interactive image exploration.

An IGA incorporating the proposed Diffusion crossover
was implemented,
and image exploration was conducted on both the MNIST and ModelNet datasets.
For MNIST, the target concept was assumed to be
a ``thick and bold 5'',
while for ModelNet,
a ``cushioned L-shaped sofa'' was assumed.
The authors acted as users
and selected images interactively
to observe the progression of the search.

At each generation, the interaction proceeded as follows:
\begin{enumerate}
  \item $N=10$ images were generated using the DDPM,
        and the corresponding noise sequences were stored.
  \item The user selected two preferred images.
  \item The corresponding noise sequences were interpolated via Slerp,
        and $N$ new image candidates were generated
        through the reverse diffusion process.
\end{enumerate}

Through this experiment, we confirmed that the proposed method
effectively functions
as a human-in-the-loop evolutionary image exploration framework.


\section{Results and Discussion}
\subsection{Results of Experiment 1: Validation of Diffusion crossover Behavior}
\begin{figure}[htbp]
    \centering
    \includegraphics[
width=\linewidth,
bb=00 00 800 450,
  clip
    ]{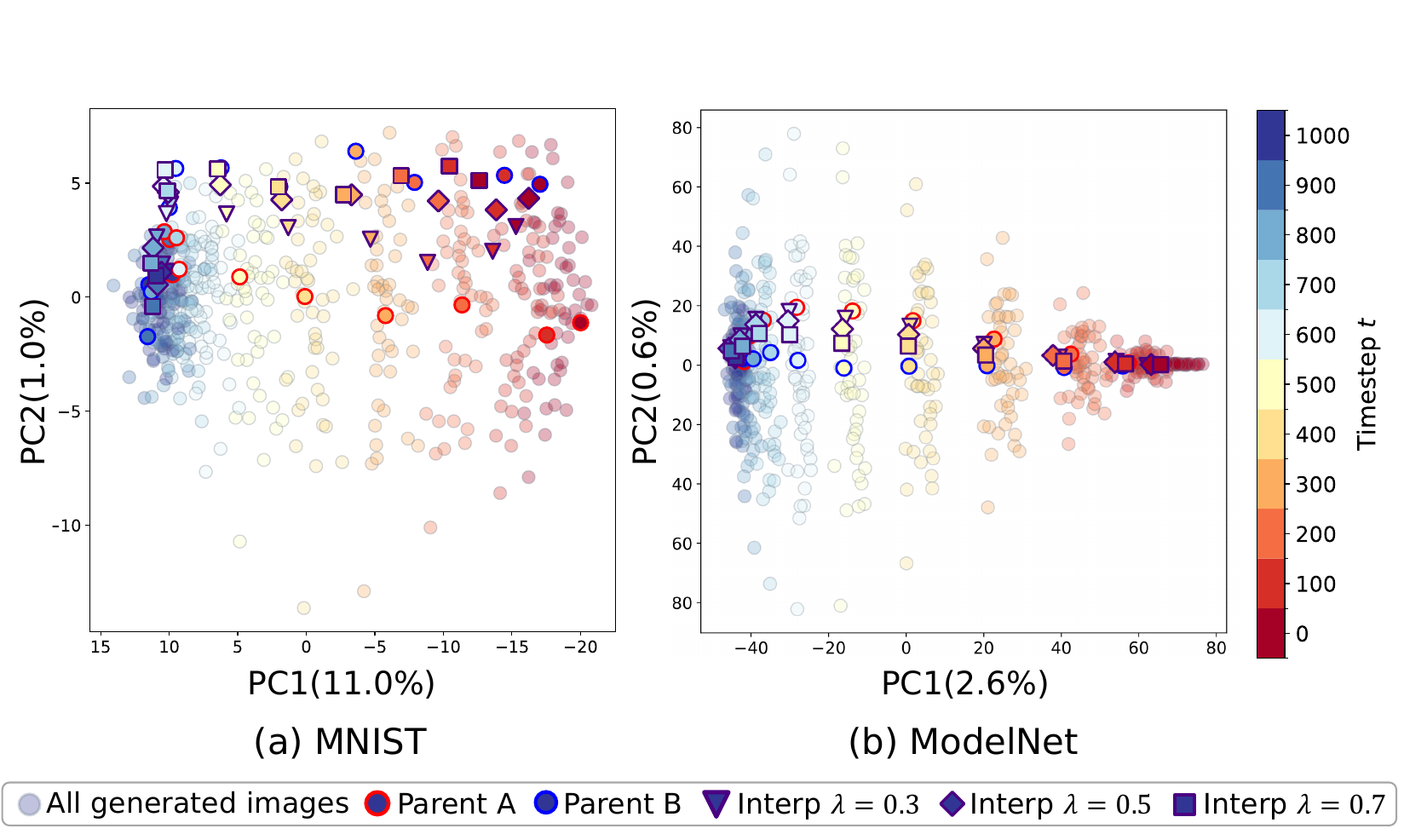}
    \caption{PCA visualization of intermediate images during reverse diffusion.
    Each point represents an intermediate image, and colors indicate time steps.}
    \label{img:pca_interp}
\end{figure}

Fig.~\ref{img:pca_interp} shows the PCA results
for intermediate images in the generation process
as well as images generated using the proposed interpolated noise.
In the figure, points corresponding to intermediate images are color-coded by time step.

The analysis reveals that, at each time step, the images generated from interpolated noise
are distributed at intermediate positions
between the corresponding parent trajectories,
and their characteristics vary continuously
with the interpolation coefficient.
This suggests that the proposed method
enables semantically meaningful interpolation in the noise space
while preserving the geometric structure
of the diffusion process.

\subsection{Results of Experiment 2: Continuity Induced by Noise Sequence Interpolation}

\begin{figure}[htbp]
    \centering
    \includegraphics[
width=\linewidth,
bb=00 00 500 500,
  clip
    ]{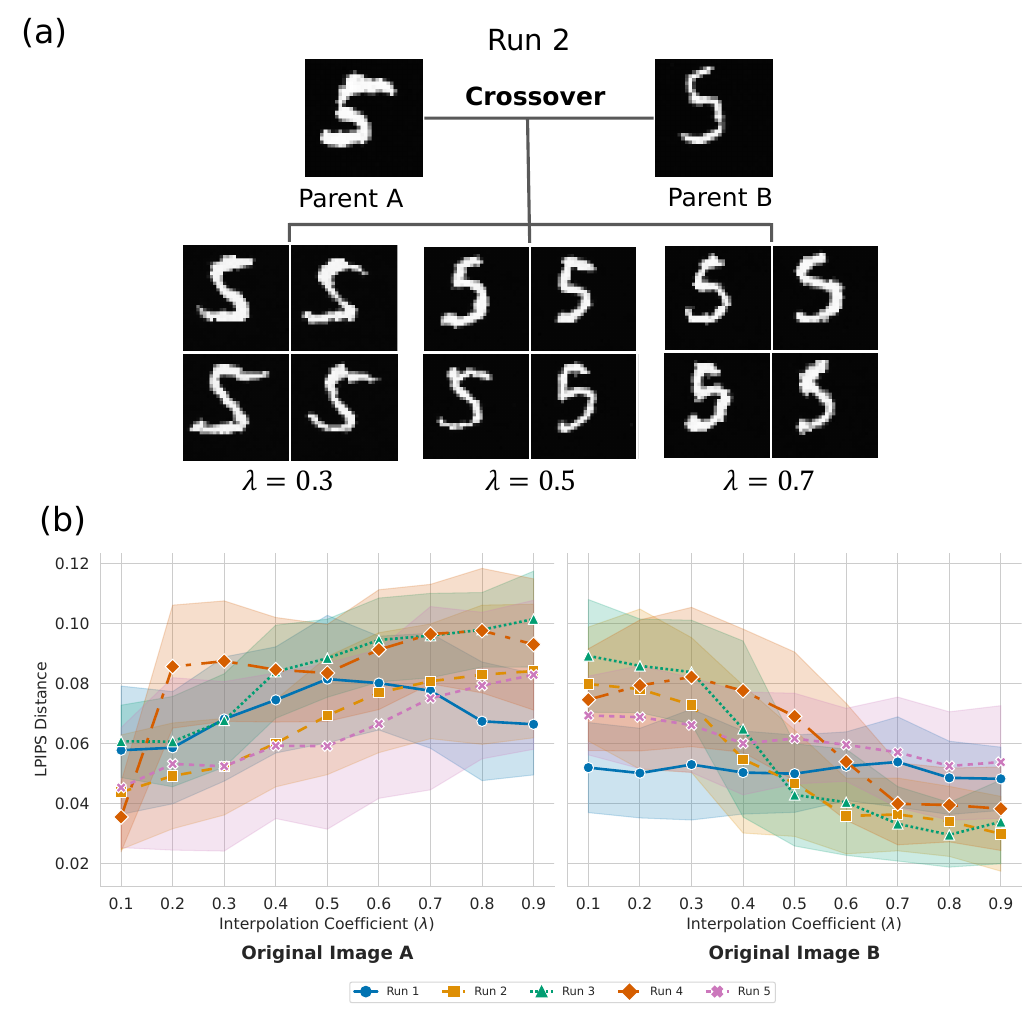}
    \caption{(a) Generated images with varying interpolation coefficients $\lambda$ (selected from the run with the highest correlation). (b) Relationship between the interpolation coefficient $\lambda$
    and LPIPS on MNIST.}
    \label{img:mnist_lpips_vs_lam}
\end{figure}

\begin{figure}[htbp]
    \centering
\includegraphics[
width=1.\linewidth,
bb=00 00 500 470,
  clip
]{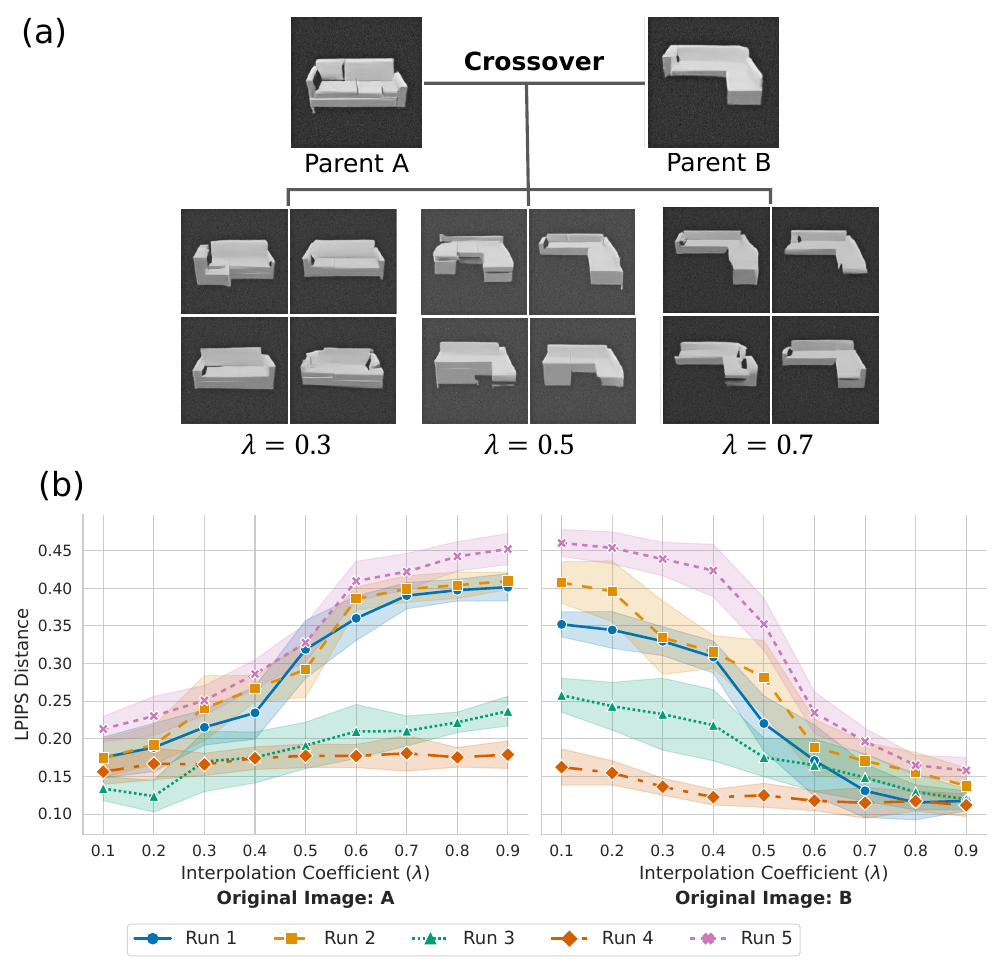}
    \caption{(a) Generated images with varying interpolation coefficients $\lambda$ (selected from the run with the highest correlation). (b) Relationship between the interpolation coefficient $\lambda$
    and LPIPS on ModelNet (Sofa).}
    \label{img:sofa_lpips_vs_lam}
\end{figure}

Fig.~\ref{img:mnist_lpips_vs_lam} and Fig.~\ref{img:sofa_lpips_vs_lam} show the generated images and LPIPS trends
for the MNIST and the ModelNet dataset, respectively, as the interpolation coefficient $\lambda$ varies.
These results demonstrate that the proposed noise sequence interpolation
functions as a crossover operator in evolutionary computation.

From a qualitative perspective, as shown in Fig.4 and Fig.5,
as $\lambda$ increases, the characteristics of the generated images
exhibit a progressive transition from parent $\mathbf{x}^{\rm (A)}$ to parent $\mathbf{x}^{\rm (B)}$.
Even for intermediate $\lambda$, the generated images do not collapse,
and the geometric consistency of the object—e.g., the sofa structure—
is preserved.

Quantitative evaluations support this observation.
As shown in Fig.6 and Fig.7,
the LPIPS distance between generated images and $\mathbf{x}^{\rm (A)}$ increases monotonically with $\lambda$,
while the distance to $\mathbf{x}^{\rm (B)}$ decreases.
This monotonic trend is statistically validated by significant rank correlations ($p<0.01$) across all independent runs for both datasets,
with the sole exception of Run 1 in the MNIST dataset (Table~\ref{tab:all_results}).
Collectively, the proposed noise sequence interpolation achieves a monotonic transition in perceptual distance space, effectively controlling the semantic mixing ratio in a predictable manner.

It is worth noting that LPIPS variation for MNIST is relatively high compared to ModelNet, and the trend is less distinct in some runs.
This can be attributed to the domain gap between MNIST (handwritten digits) and the natural images used to train LPIPS, which can limit the metric’s sensitivity.
However, given the robust results in the majority of MNIST runs and ModelNet, the overall effectiveness of the proposed method is confirmed.

\begin{table}[t]
\centering
\caption{Spearman's rank correlation coefficients ($\rho$) for Experiments 2 and 3.}
\label{tab:all_results}
\begin{tabular}{@{}lccccc@{\quad}}
\toprule
                         & Run1                 & Run2                 & Run3                 & Run4                 & Run5                 \\ \midrule
\multicolumn{6}{@{}l}{\textbf{Experiment 2 ($\bm \lambda$ vs. LPIPS)}}                                                                              \\
~~MNIST (Ref. A)    & \phantom{-}0.33                 & \phantom{-}1.00\rlap{\textsuperscript{***}}               & \phantom{-}0.98\rlap{\textsuperscript{***}}              & \phantom{-}0.80\rlap{\textsuperscript{**}}               & \phantom{-}0.97\rlap{\textsuperscript{***}}              \\
~~MNIST (Ref. B)    & -0.37                & -0.98\rlap{\textsuperscript{***}}             & -0.95\rlap{\textsuperscript{***}}             & -0.88\rlap{\textsuperscript{**}}              & -0.97\rlap{\textsuperscript{***}}             \\
~~ModelNet (Ref. A) & \phantom{-}1.00\rlap{\textsuperscript{***}}               & \phantom{-}1.00\rlap{\textsuperscript{***}}               & \phantom{-}0.98\rlap{\textsuperscript{***}}              & \phantom{-}0.83\rlap{\textsuperscript{**}}               & \phantom{-}1.00\rlap{\textsuperscript{***}}               \\
~~ModelNet (Ref. B) & -0.98\rlap{\textsuperscript{***}}             & -1.00\rlap{\textsuperscript{***}}              & -1.00\rlap{\textsuperscript{***}}              & -0.96\rlap{\textsuperscript{***}}             & -1.00\rlap{\textsuperscript{***}}              \\
\multicolumn{6}{@{}l}{\textbf{Experiment 3 ($\bm{t_{\rm interp}}$ vs. diversity score)}}                                                                      \\
~~MNIST                    & -0.98\rlap{\textsuperscript{***}}             & -0.95\rlap{\textsuperscript{***}}             & -0.83\rlap{\textsuperscript{**}}              & -0.93\rlap{\textsuperscript{***}}             & -0.98\rlap{\textsuperscript{***}}             \\
~~ModelNet                 & -0.98\rlap{\textsuperscript{***}} & -0.95\rlap{\textsuperscript{***}} & -0.98\rlap{\textsuperscript{***}} & -0.93\rlap{\textsuperscript{***}} &
-0.6
\\ \bottomrule
\multicolumn{6}{l}{\small $^{*} p < 0.05,\ ^{**} p < 0.01,\ ^{***} p < 0.001$}                                                         
\end{tabular}
\end{table}

\subsection{Results of Experiment 3: Effect of Interpolation Duration on Diversity}
\begin{figure}[htbp]
    \centering
    \includegraphics[
width=\linewidth,
bb=00 00 750 400,
  clip
  ]{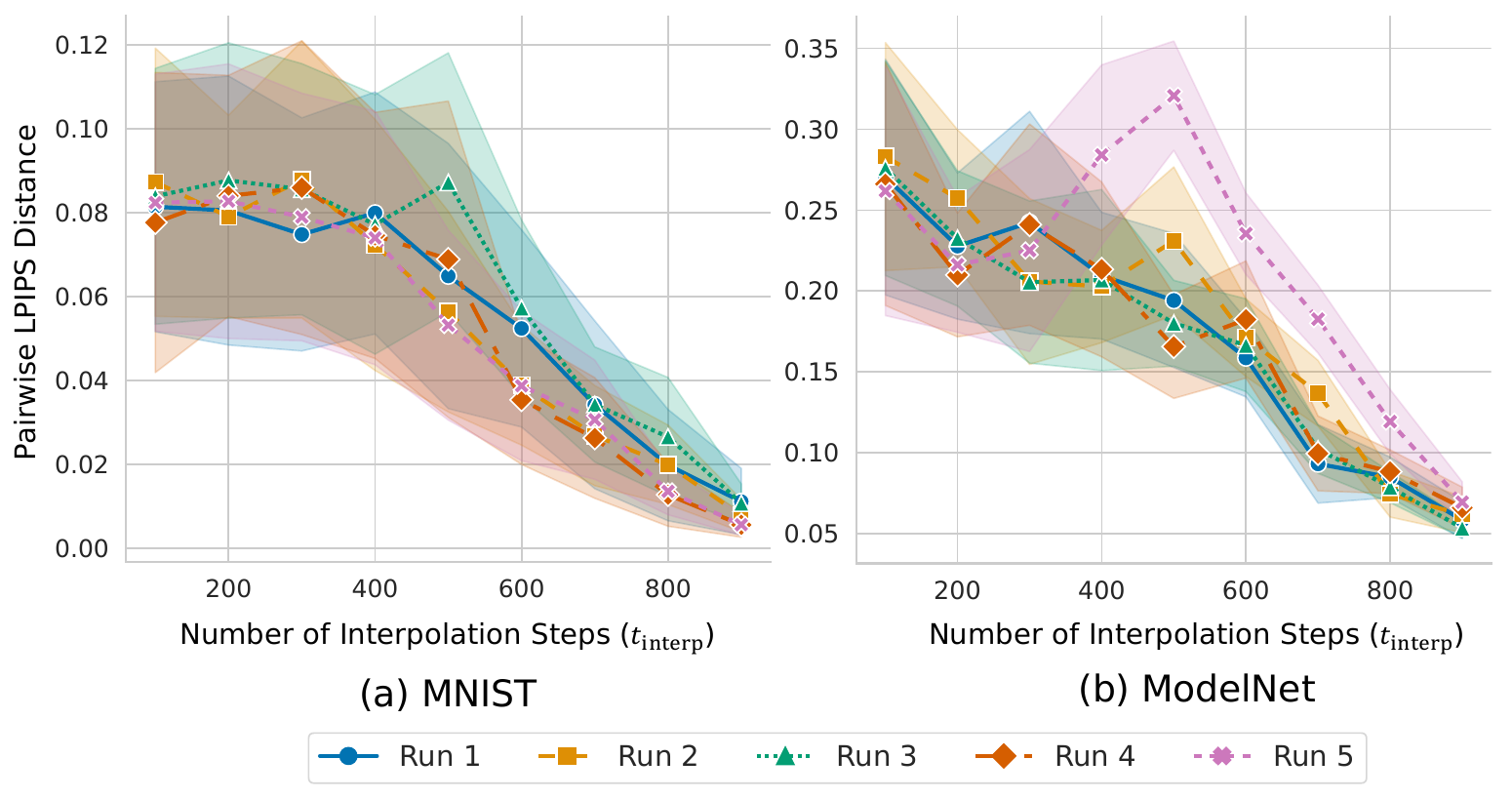}
    \caption{Relationship between $t_{\rm interp}$ and diversity score on (a) MNIST and (b) ModelNet.}
    \label{img:lpips_vs_t}
\end{figure}

Fig.~\ref{img:lpips_vs_t} shows the relationship between
the interpolation duration $t_{\rm interp}$
and the diversity score of generated images (average pairwise LPIPS).
For both datasets, as $t_{\rm interp}$ increases, the diversity score decreases monotonically.

This trend is statistically validated by the Spearman rank correlation analysis (Table~\ref{tab:all_results}).
Significant negative correlation was demonstrated
in all runs for MNIST ($p<0.05$) and all but one run for ModelNet ($p<0.01$).

These results can be interpreted as follows: as $t_{\rm interp}$ increases, the duration in which random noise is newly sampled during generation becomes shorter, and the outputs are more strongly constrained
by the interpolated noise trajectory.

This result suggests that,
by controlling the interpolation duration $t_{\rm interp}$,
the balance between exploration (diversity)
and exploitation (convergence)
can be adjusted.

\subsection{Results of Experiment 4: Qualitative Results of the Interactive Evolutionary Process}
\begin{figure}[htbp]
    \centering
    \includegraphics[
width=\linewidth,
bb=00 00 800 410,
  clip
    ]{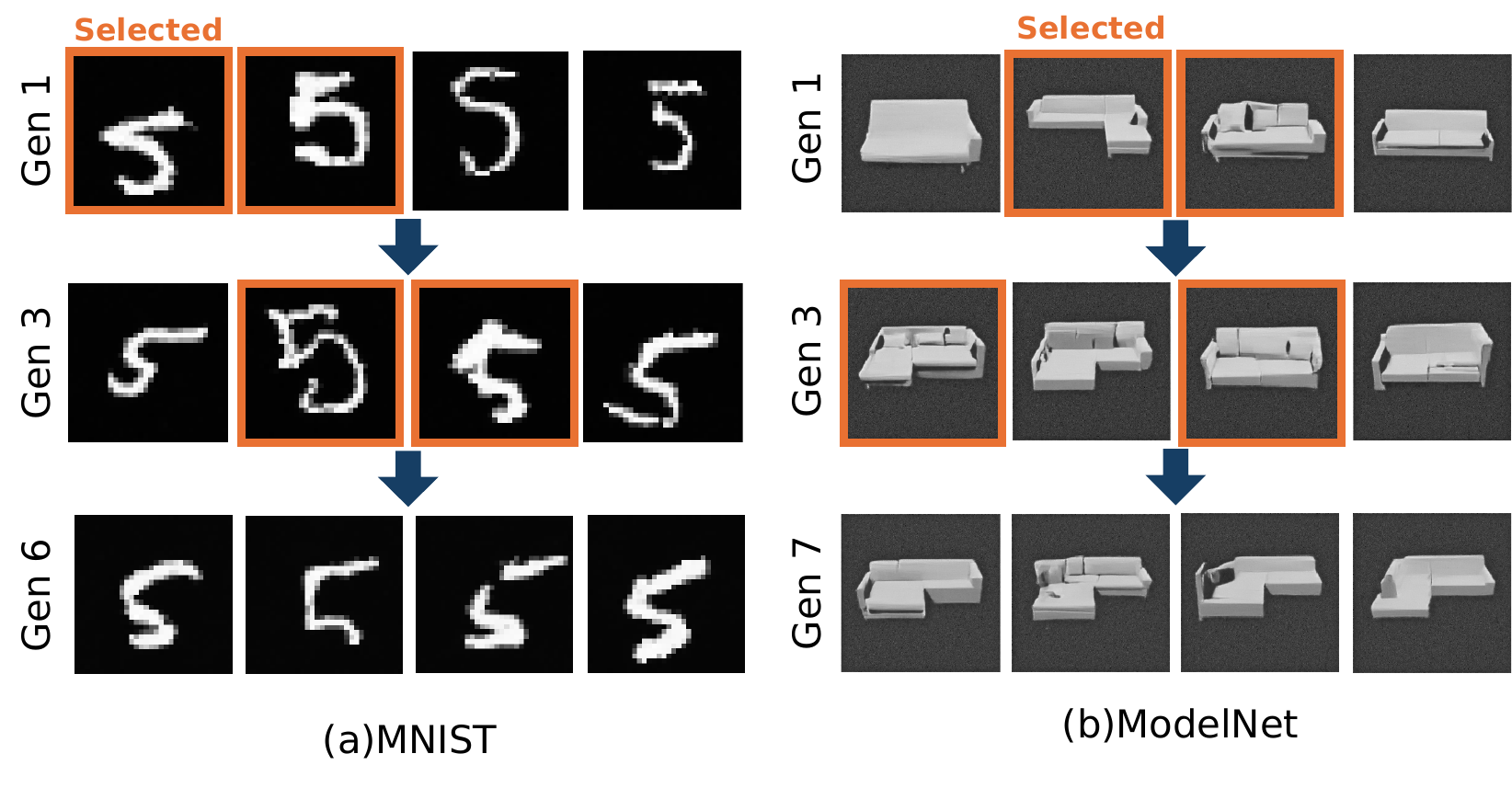}
    \caption{Examples of the interactive evolutionary process.
    (a) MNIST results. (b) ModelNet (sofa) results.}
    \label{img:iga_all}
\end{figure}

Fig.~\ref{img:iga_all} presents qualitative results
of the interactive evolutionary process using the proposed method.
By repeatedly selecting images based on user preference at each generation, the generated images gradually converge
toward specific shapes and visual characteristics.

In the initial generation (Gen 1), the population exhibits high diversity.
However, through iterative selection and the proposed crossover,
phenotypic traits align progressively with the user’s intent.
For the MNIST task targeting a thick and bold 5'', while the initial population contains various digits and styles, the population at Gen 6 is dominated by digits with desired thick strokes. In the ModelNet task aiming for a cushioned L-shaped sofa, the generation process evolves from diverse shapes in Gen 1
to consistent L-shaped structures by Gen 7.
It was observed that while global geometric features were robustly inherited across generations, fine-grained textures (e.g., cushioning details) showed greater variability.
This discrepancy can be attributed to the inherent coarse-to-refine nature of the DDPM generative process.
Because the noise interpolation is performed during the early denoising steps,
it dominantly influences the global structure.
Developing methods to achieve fine-grained control over such localized features requires further investigation.

These results indicate that the proposed Diffusion crossover
is not merely a numerical interpolation technique,
but an operator that can effectively function
within human-in-the-loop evolutionary exploration.

\subsection{General Discussion}
\paragraph{Computational Cost and Interactivity}
A practical concern in applying diffusion models to IEC
is the computational cost of sampling
and the resulting latency in user interaction.
In our implementation, generating a single population of images
requires approximately 40–50 seconds on a GPU.
This latency stems from the $T=1000$ iterative denoising steps of DDPM and from the computational overhead of saving and applying Slerp to high-dimensional noise sequences used in the reverse process.

While this waiting time poses a challenge for real-time user interaction, the primary focus of this study is to validate the fundamental capability of the proposed noise-sequence interpolation as a crossover operator.

For future practical application, this latency can be reduced.
While accelerated sampling methods were not employed in this study to evaluate the fundamental behavior of the standard DDPM noise sequence,
it is, in principle, compatible with accelerated sampling methods that reduce the number of DDPM steps.
Even with fewer sampling steps, the interpolation operation in the noise space can be defined in the same manner.
In addition, the proposed interpolation method is theoretically compatible with Latent Diffusion Models (LDMs)~\cite{rombach2022highresolution}.
Because LDMs operate in a compressed latent space,
applying our method to LDMs is expected to reduce the dimensionality and computational cost of noise sequences.
Thus, extending the framework to settings
that prioritize responsiveness is a promising direction.

\paragraph{Linearity and Non-spatiality of Noise Interpolation}
We hypothesize that a naive spatial crossover (e.g., splicing noise sequences) would be ineffective in this framework.
Although not explicitly tested, the behavior observed during our linear interpolation strongly supports this hypothesis.
For instance, using an L-shaped sofa as a parent occasionally yielded flipped L-shapes or U-shaped structures.
This implies that the spatial coordinates of the noise sequence do not map to the generated image structure in a localized manner.
Rather, as the noise is processed by the U-Net architecture, it may acquire higher-dimensional representations.
This highlights the need to apply global operations rather than naive spatial splicing to achieve semantic integration.

\paragraph{Influence of Initial vs. Added Noise}
While previous studies have shown that spatially editing the initial noise can partially alter generated images, our observations indicate that the noise added during subsequent denoising steps contributes more significantly to the image generation.
This likely arises because existing spatial edits rely on conditional generation and deterministic samplers.
In our unconditional DDPM framework, continuous stochastic noise injection heavily dictates the generation trajectory,
making sequence-wide interpolation essential.

\paragraph{Limitations of the Qualitative User Evaluation}
The interactive evolutionary experiment in this study
was conducted as a feasibility study
to confirm that the proposed method
can function in a human-in-the-loop exploration setting.
Accordingly, the authors acted as users
and qualitatively evaluated the generation behavior.

A third-party user study, even at a small scale,
and statistical evaluation based on metrics such as user satisfaction and convergence speed would be important for more rigorously validating practicality, and we leave this as future work.

\paragraph{Model Selection and Hyperparameter Validity}
In this study, model selection was based on generation quality, as measured by CMMD.
However, in the IEC context, the fidelity of a generative model does not necessarily align
with search efficiency or user satisfaction.

A more comprehensive hyperparameter optimization strategy that accounts for exploration diversity and navigability, along with generation quality, is an important topic for future research.

\section{Conclusion}
This study proposed \emph{Diffusion crossover}, a novel crossover operation for Interactive Evolutionary Computation (IEC) explicitly defined as the interpolation of noise sequences during the generation process of Denoising Diffusion Probabilistic Models (DDPM).
By integrating noise sequences from multiple generated samples, the proposed method enables offspring to inherit characteristics from parent individuals, extending diffusion-based evolutionary search beyond mutation-dominated operations.

Through numerical experiments, PCA and LPIPS-based evaluations demonstrated that the proposed method preserves the geometric structure of the diffusion process while achieving perceptually continuous image transitions.
In addition, controlling the interpolation duration was shown to effectively balance diversity and convergence in the generated images.

Furthermore, qualitative evaluations of an interactive evolution confirmed that Diffusion crossover functions effectively in a human-in-the-loop setting, enabling gradual, preference-driven convergence.
These results indicate that the proposed framework provides a practical and general approach for utilizing diffusion models not merely as generators, but as operative search spaces for evolutionary exploration.


\newpage
\printbibliography

\end{document}